\documentclass[runningheads]{llncs}
\usepackage[T1]{fontenc}
\usepackage{booktabs} 
\usepackage{graphicx}
\usepackage{amsmath}
\begin{document}
\title{Interpretable Human Activity Recognition for Subtle Robbery Detection in Surveillance Videos}
\titlerunning{Robbery Detection in Surveillance Videos}
\author{Bryan Jhoan Cazáres Leyva\thanks{These authors contributed equally to this work.}\inst{1} \and
Ulises Gachuz Davila$^{\star}$\inst{1} \and
José Juan González Fonseca$^{\star}$\inst{1} \and Juan Irving Vasquez\inst{2}\orcidID{0000-0001-8427-9333} \and Vanessa A. Camacho-Vázquez\inst{1}\orcidID{0000-0002-1149-3950} \and Sergio Isahí Garrido-Castañeda \inst{2}\orcidID{0009-0003-4549-7794}}
\authorrunning{Cázares et al.}
\institute{Escuela Superior de Cómputo, Instituto Politécnico Nacional, México City, México. \and
Centro de Innovación y Desarollo Tecnológico en Cómputo, Instituto Politécnico Nacional, México City, México.\\
\email{jvasquezg@ipn.mx}}
\maketitle              % typeset the header of the contribution
\begin{abstract}
Non-violent street robberies (snatch-and-run) are difficult to detect automatically because they are brief, subtle, and often indistinguishable from benign human interactions in unconstrained surveillance footage. This paper presents a hybrid, pose-driven approach for detecting snatch-and-run events that combines real-time perception with an interpretable classification stage suitable for edge deployment. The system uses a YOLO-based pose estimator to extract body keypoints for each tracked person and computes kinematic and interaction features describing hand speed, arm extension, proximity, and relative motion between an aggressor-victim pair. A Random Forest classifier is trained on these descriptors, and a temporal hysteresis filter is applied to stabilize frame-level predictions and reduce spurious alarms. We evaluate the method on a staged dataset and on a disjoint test set collected from internet videos, demonstrating promising generalization across different scenes and camera viewpoints. Finally, we implement the complete pipeline on an NVIDIA Jetson Nano and report real-time performance, supporting the feasibility of proactive, on-device robbery detection.

\end{abstract}
\section{Introduction}

Public safety, particularly in regard to non-violent street robbery, remains a critical challenge for urban security \cite{Tykesson2025}. Despite the global expansion of video surveillance infrastructure, its effectiveness is often hindered by heavy reliance on real-time human interpretation. 

\begin{figure}[htb]
    \centering
    \includegraphics[width=0.80\linewidth]{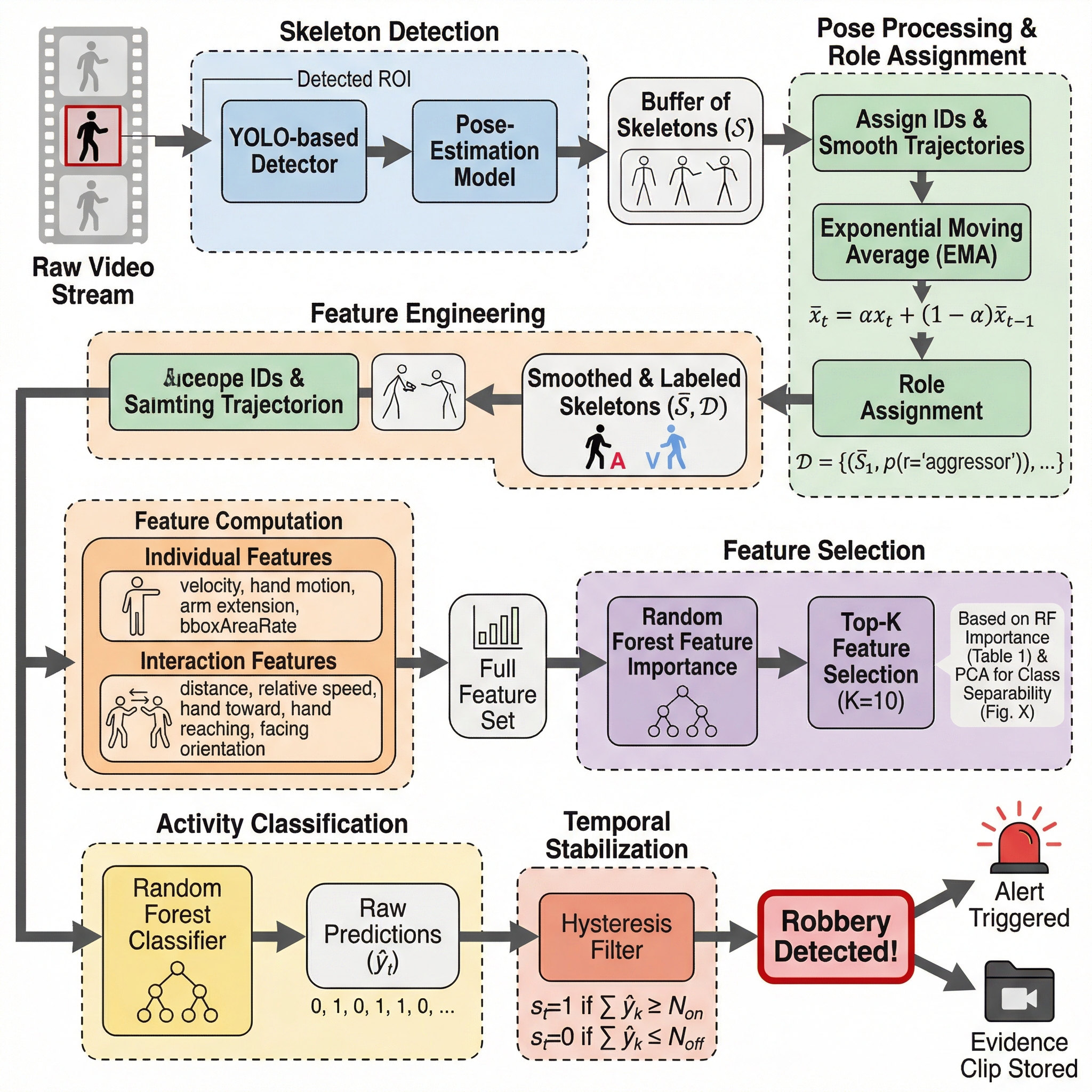}
    \caption{General flow of the proposed method for the recognition of robbery in surveillance videos.}
    \label{fig:generalflow}
\end{figure}

%The recognition of the mentioned human interactions, such as robbery or the act of taking an object from another person, has become an important research topic in computer vision due to its importance in intelligent surveillance and robotics \cite{ren2024survey}. 

Traditionally, action recognition relied on computationally intensive modalities like RGB image sequences or depth videos \cite{liu2019ntu}. However, 3-dimensional (3D) skeleton-based action recognition (SAR) has emerged as a more robust alternative, providing detailed topological representations of the human body through joints and bones \cite{liu2024transtm}. Early methods primarily utilized handcrafted features to capture relative 3D rotations and translations among body parts \cite{poppe2010survey}. Recent advancements have shifted toward deep learning architectures, specifically Recurrent Neural Networks (RNNs) and Long Short-Term Memory (LSTM) networks, which excel at capturing temporal dependencies in joint movements \cite{shu2020host}\cite{balakrishnan2024accurate}. Despite their temporal proficiency, standard RNNs often lack spatial modeling capabilities \cite{wang2023comprehensive}. To address this, Convolutional Neural Networks (CNNs) are frequently employed to identify spatial features and local patterns by transforming skeleton sequences into pseudo-image formats \cite{ko2018deep}. Furthermore, Graph Convolutional Networks (GCNs) have gained prominence by treating the human skeleton as a natural graph structure, effectively modeling the interdependence between joints and bones \cite{liu2020disentangling}. For high-stakes interactions like robbery, modern approaches increasingly leverage Transformer-based architectures and hybrid models to capture long-range dependencies and global relationships within the data \cite{wang20233mformer,liu2024transtm}. These methodologies are typically evaluated on large-scale datasets such as NTU-RGB+D and NTU-RGB+D 120, which offer challenging cross-subject and cross-view evaluation protocols \cite{liu2019ntu}. While existing studies explore violent behavior detection using pose estimation and neural networks, these models target broad categories of aggression. Consequently, there remains a lack of specialized systems focusing on the specific subtle kinematics of non-violent robbery incidents between two individuals.
% previous work
% Se quito para poner aqui la sección de related work
%Prior work on human interaction understanding has progressed from computationally heavy RGB/RGB-D pipelines to skeleton-based action recognition, which offers greater robustness to viewpoint and illumination changes \cite{hu2015jointly,liu2019ntu,yan2018spatial}. Recent methods primarily rely on deep sequence and graph models (e.g., RNN/LSTM, CNN-based encodings, GCNs, and Transformers) to capture spatiotemporal dynamics of body joints \cite{shu2020host,liu2020disentangling,wang20233mformer}. This literature motivates the use of pose keypoints and kinematic descriptors as an efficient representation for detecting subtle, high-stakes events such as snatch-and-run robberies \cite{ren2024survey,liu2024transtm}. 

%Furthermore, the high operational costs and proprietary nature of existing high-end surveillance platforms limit their accessibility for public safety deployment. There is a critical need for a cost-effective, proactive system capable of transitioning from simple object detection to human intent analysis through pose estimation, optimized for execution on edge computing devices.

Although AI-driven video analytics have significantly reduced false alarms and improved detection of defined actions such as shoplifting or explicit violent behaviors \cite{Mohammadi2022}, research and technical deployments indicate substantial gaps remain in automating the recognition of subtle, non-violent behavioral signatures (e.g., rapid snatching followed by immediate flight), which are harder to characterize and detect using current models. Furthermore, end-to-end methods based on deep learning lack of explainability.

% proposal
This paper addresses the detection of non-violent robbery (snatch-and-run) by proposing a method that combines neural-network-based perception with an explainable, feature-based classifier. The system uses the YOLO architecture to detect individuals and estimate their body keypoints. Next, an interpretable feature-extraction stage computes kinematic and interaction descriptors from the skeleton trajectories; a formally ranked subset of the most relevant features is then used to detect the event using a Random Forest classifier. The main contribution of this work is the proposed explainable classification pipeline for pose-driven robbery recognition.

%alcances
The implemented system is evaluated in controlled environments designed to replicate real-world theft dynamics. We report accuracy and response time to establish a baseline for proactive surveillance efficiency relative to traditional manual monitoring. In addition, the experiments show that the method can run on the NVIDIA Jetson Nano, supporting its suitability for edge deployment.

% Quitado para disminuir espacio
%The remainder of this paper is organized as follows. Section \ref{sec:rl} reviews related work on interaction and skeleton-based action recognition. Section \ref{sec:recognition} describes the proposed robbery-recognition pipeline, including pose extraction, feature computation, and temporal filtering. Section \ref{sec:experiments} details the experimental setup, datasets, and reports quantitative results and discusses the findings. Finally, Section \ref{sec:conclu} concludes the paper and outlines future work.

%\section{Related Work}
%\label{sec:rl}

\section{Robbery Recognition Methodology}
\label{sec:recognition}

% Primer párrafo describe el método de forma general
Overall, the proposed method takes a raw video stream as input. First, a YOLO-based detector identifies people in each frame; then, for each detected ROI, a pose-estimation model computes the corresponding skeleton. Next, we assign consistent IDs to the detected individuals and smooth the resulting pose trajectories over time. For each person, we refine the pose sequence by computing relative features and then retain only a subset of the most informative ones. These selected features are fed into a previously trained Random Forest classifier to predict the activity class. Because the classifier may produce sporadic misclassifications, we apply a hysteresis-based decision rule to stabilize predictions over time. Finally, when a robbery is detected, the system triggers an alert and stores the corresponding evidence. A general flow diagram is presented in Fig. \ref{fig:generalflow}, and the details are provided next.

\subsection{Skeleton detection}

The raw video is split into a frame buffer. Skeletons are then extracted in two stages: first, the YOLO detector identifies the people in the scene; next, for each detected bounding box, a pose-estimation model predicts the corresponding skeleton. This produces a buffer of skeletons over time, denoted as $\mathcal{S} = \{S_1,\ldots,S_n\}$, where $n$ is the number of people in the scene. Each skeleton consists of 17 body keypoints: the nose (0), eyes (1,2), ears(3, 4), shoulders (5, 6), elbows (7, 8), wrists (9, 10), hips (11, 12), knees (13, 14), and ankles (15,16).

\subsection{Pose estimation}

To reduce noise in the skeleton measurements over time, $\mathcal{S}$, we smooth each joint trajectory using an Exponential Moving Average (EMA). For a joint coordinate $x_t$ at time $t$, the EMA is defined as shown in Eq. (\ref{eq:ema}):\begin{equation}\bar{x}t = \alpha x_t + (1-\alpha)\bar{x}{t-1}, \qquad 0<\alpha<1,\label{eq:ema}\end{equation}with initialization $\bar{x}_0 = x_0$. Applying this update to every joint and coordinate yields the smoothed skeleton sequence $\bar{\mathcal{S}}$.

In this human-to-human interaction scenario, there are two possible roles: victim and aggressor. Because we have no prior knowledge of each person’s intent, we evaluate both role assignments and attach a probability to each one. Thus, we augment the set of detected and filtered skeletons with an aggressor label, represented by the set $\mathcal{D}$ in Eq. (\ref{eq:aggressor_set}).

\begin{equation}\mathcal{D} = {(\bar{S}_1, p(r=\text{'aggressor'})), \ldots, (\bar{S}_n, p(r=\text{'aggressor'}))},\label{eq:aggressor_set}\end{equation}
where the probability of being an aggressor is estimated from the mean translation of the skeleton over time and converted into a normalized score using a softmax function. This heuristic assumes that the aggressor (or thief) exhibits more abrupt movements.

\subsection{Explainable feature extraction}

This section describes the interpretable kinematic and interaction features computed from the smoothed skeleton sequences. All distances and velocities are normalized by the torso height to reduce sensitivity to scale changes caused by subject--camera distance. We group the features into two categories: 

\subsubsection{Individual features per track.}
They characterize the motion and posture of each person independently. They are divided in four groups:

%In the following, the prefix $[A/B]$ indicates that the same feature is computed for person A and person B.

\begin{itemize}
    \item \textbf{Body-center kinematics features.}
    $\texttt{velocity}$: normalized velocity of the person center. $\texttt{acceleration}$:  normalized acceleration of the person center.
    \item \textbf{Hand motion features.} \texttt{handVelocity}: normalized wrist speed. High values suggest rapid hand movement (e.g., a slap, grab, or snatch attempt).
    \texttt{fastHandPct}: percentage of frames for which the wrist speed exceeds a fixed threshold. It measures the duration of fast hand motion.
    \texttt{timeToPeakHandVel}: number of frames required to reach the maximum wrist speed within the segment (timing of the fastest motion).
    $\texttt{handAcceleration}$:('aggressor') wrist acceleration.
    $\texttt{handJerkMin}$:('aggressor') minimum wrist jerk (derivative of acceleration).
    \item \textbf{Arm extension and posture features.}
    \texttt{armExtension}: normalized arm extension (wrist-shoulder distance). $\texttt{timeToPeakArmExt}$('aggressor'): frames until it reaches the max. arm extension. $\texttt{armRetraction0p2s}$('aggressor'): drop in aggressor's arm extension $0.2\,\mathrm{s}$ after its peak. This ``retraction'' metric captures whether the aggressor quickly withdraws the arm after reaching maximum reach. \texttt{elbowFlexPct}[L/R]: percentage of frames in which the elbow angle is below a threshold indicating a tendency to keep the arms flexed. $\texttt{elbowAngle}[L/R]$: Estimation of the elbow angle.
    \item \textbf{Bounding-box features.} $\texttt{bboxAreaRate}$: relative derivative of the bounding-box area. Positive values may indicate motion toward the camera.
\end{itemize}

\subsubsection{Interaction features.} Characterize the joint movement. Subjects A and B.
\begin{itemize}
    \item \textbf{Distance and contact features:}  $\texttt{distance}$:  normalized distance between the centers of subjects.
        $\texttt{distanceRate}$: distance change rate (approach/separation speed). 
        $\texttt{iou}$: IoU between their bounding boxes (physical contact or crowding).
        $\texttt{iouPeak}$: maximum IoU within the segment.
        $\texttt{iouDrop0p2s}$: IoU drop $0.2\,\mathrm{s}$ after the peak, measuring how quickly they separate after maximum overlap.
    \item \textbf{Relative movement features.}
    $\texttt{relativeSpeed}$: relative speed between the centers of subjects.
    $\texttt{handTowardCos}$: cosine similarity between A's hand velocity vector and the vector from A to B's torso. Values close to 1 indicate that A's hand moves directly toward B, serving as an intent cue.
    $\texttt{handTowardPct}$: percentage of frames with high directionality, measuring the duration of ``attack intent.''
    \item \textbf{Hand reaching features.}
    $\texttt{handToTorsoMin}$: minimum normalized distance from A's wrist to B's torso center.
    $\texttt{closeHandPct}$: percentage of frames in which agrressor's hand is very close to victims's torso.
    $\texttt{handToHipMin}$: minimum normalized distance from aggressors's wrist to victims's hips, which can help detect interactions directed to the lower torso (e.g., pockets).
    $\texttt{fastAndClosePct}$: percentage overlap of ``fast hand'' and ``close hand'' conditions, combining rapid action with close interaction.
    $\texttt{fastAndCloseLongest}$: longest consecutive run (frames) satisfying the ``fast and close'' condition.
    $\texttt{postContactSepMean}$: mean center distance in a $0.4\,\mathrm{s}$ window after A's hand reaches its closest point to B's torso, capturing immediate withdrawal/separation.
    \item \textbf{Relative face orientation features.}
    $\texttt{BfacingToA}$: cosine similarity between B's facing direction and the vector from B to A. Values close to $-1$ indicate that B is facing A, while values close to $1$ indicate that B is turned away.
    $\texttt{AfacingToB}$: cosine similarity between A's facing direction and the vector from A to B. Values close to $1$ indicate that A is facing B.
    $\texttt{facingRate}$: derivative of the facing direction.
\end{itemize}

\subsection{Feature Selection Analysis}

To select a compact and informative subset of variables, we first trained a Random Forest model using the full set of available features. We then used the model's feature importances attribute to estimate the contribution of each feature to the classifier's decision process. Based on this ranking, we retained the 10 most important features, which helps reduce noise and redundancy while preserving discriminative power. The choice of 10 features was made empirically.

Table~\ref{tab:feat_importance} lists the selected features in descending order of importance. Among the top-10, $\texttt{AB\_dist\_p95}$ provides the highest contribution to classification, whereas $\texttt{AB\_handTowardGt07Pct}$ has the lowest contribution within the selected subset.

\begin{table}[ht]
\centering
\caption{Top 10 features ranked by Random Forest importance (Two columns).}
\label{tab:feat_importance}
\begin{tabular}{|c|l|c||c|l|c|}
\hline
\textbf{Rank} & \textbf{Feature} & \textbf{Imp. ($10^{-2}$)} & \textbf{Rank} & \textbf{Feature} & \textbf{Imp. ($10^{-2}$)} \\
\hline
1 & \texttt{dist\_p95}          & 5.236 & 6  & \texttt{distancet\_max}      & 4.188 \\
2 & \texttt{handToHip\_max}     & 4.802 & 7  & \texttt{handToTorso\_median} & 3.340 \\
3 & \texttt{handToTorso\_mean}  & 4.781 & 8  & \texttt{handToHip\_p95}      & 2.763 \\
4 & \texttt{handToTorso\_p95}   & 4.297 & 9  & \texttt{distance\_mean}      & 2.459 \\
5 & \texttt{handToTorso\_max}   & 4.239 & 10 & \texttt{closeHandPct}        & 2.352 \\
\hline
\end{tabular}
\end{table}

\begin{figure}[tb]
\centering
\includegraphics[width=0.8\linewidth]{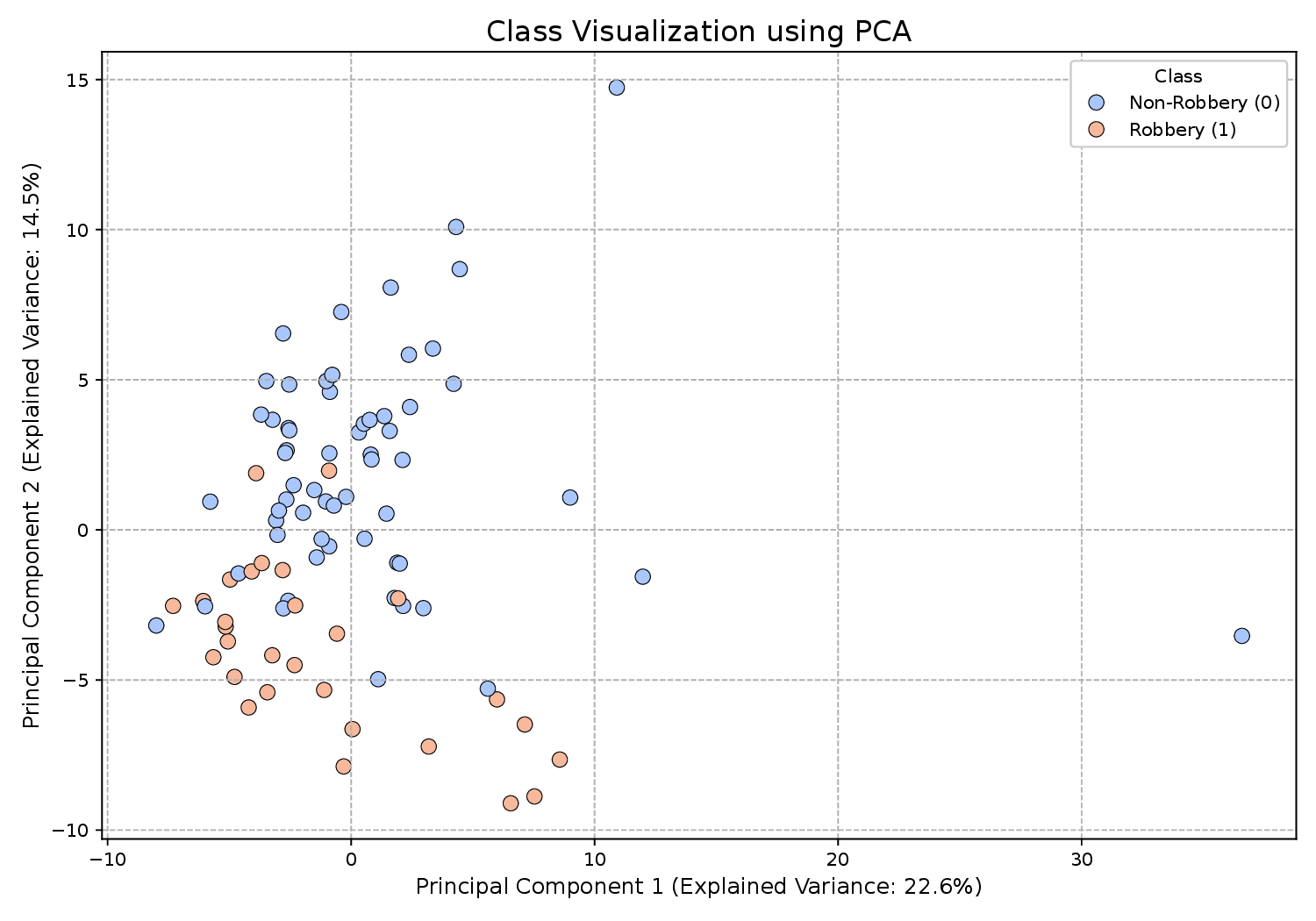}
\caption{Class visualization using PCA. As we can observe, the projected samples are in a general way distinguishable. This graph is only a projection of the feature vectors of 10 elements.}
\label{fig:figura24}
\end{figure}

Finally, to analyze class separability using the selected 10 features, we applied Principal Component Analysis (PCA) and visualized the projected samples as shown in Fig.~\ref{fig:figura24}.

\subsection{Hysteresis filtering}

Frame-by-frame predictions produced by the activity classifier may be unstable due to pose-estimation noise, partial occlusions, and short-term ambiguous motion. To avoid raising spurious alarms, we apply a temporal hysteresis filter that enforces different activation and deactivation conditions.

Let $\hat{y}_t \in \{0,1\}$ be the raw prediction at time $t$, where $\hat{y}_t=1$ denotes ``robbery''. We define an internal alarm state $s_t \in \{0,1\}$ updated as shown in Eq. (\ref{st_update}). The alarm is activated only if the classifier predicts robbery for at least $N_\mathrm{on}$ frames within a sliding window of length $W$.
\begin{equation}\label{st_update}
    s_t = 1 \quad \text{if} \quad \sum_{k=t-W+1}^{t} \hat{y}_k \ge N_\mathrm{on}
\end{equation}

Once activated, the alarm remains on until the robbery prediction has been absent for at least $N_\mathrm{off}$ frames in the same window (with $N_\mathrm{off} < N_\mathrm{on}$), a condition expressed by Eq. (\ref{eq:alarm_off}).

\begin{equation}s_t = 0 \quad \text{if} \quad \sum_{k=t-W+1}^{t} \hat{y}k \le N\mathrm{off}.\label{eq:alarm_off}
\end{equation}
Otherwise, the previous state is preserved ($s_t=s_{t-1}$). This two-threshold mechanism prevents rapid state switching and attenuates short bursts of false positives/negatives.

In practice, $W$ is chosen according to the frame rate (e.g., $W \approx 0.4\,\mathrm{s}$ of video), and $N_\mathrm{on}$ and $N_\mathrm{off}$ are tuned to trade off detection latency against robustness. When $s_t$ transitions from 0 to 1, the system triggers an alert and stores the corresponding evidence clip.

\section{Experiments}
\label{sec:experiments}

The experimentation is divided into two phases: the performance analysis of the behavioral classifier and the functional validation of the integrated prototype.

The proposed method was implemented in Python~3.12 using isolated virtual environments to ensure reproducibility. Communication between the embedded device and the desktop application is mediated by a RESTful API built with Flask, where both the NVIDIA Jetson Nano and the desktop client run independent Flask instances with complementary roles. The data flow follows the sequence Jetson Nano (detection and upload) client Flask server (storage and REST API) Electron interface (visualization and control), enabling a decoupled design.

%On the Jetson Nano, video acquisition and pre-processing are handled with OpenCV and NumPy, while human pose keypoints are estimated using Ultralytics (YOLOv11-Pose). The extracted kinematic descriptors are evaluated by a trained classifier loaded with Joblib; when an event is detected, a video clip is recorded and transmitted to the client server using the Requests library. On the desktop side, the Flask server listens on port 5500, stores metadata in a local SQLite3 database, and exposes endpoints to list, upload, delete, and retrieve video clips. Secure routing and file transfer are supported via Werkzeug, and the Electron application consumes the REST endpoints to manage and review detected events.

\begin{figure}[tb] % MDPI prefiere [H] para posición fija
    \centering
        \includegraphics[width=0.24\linewidth]{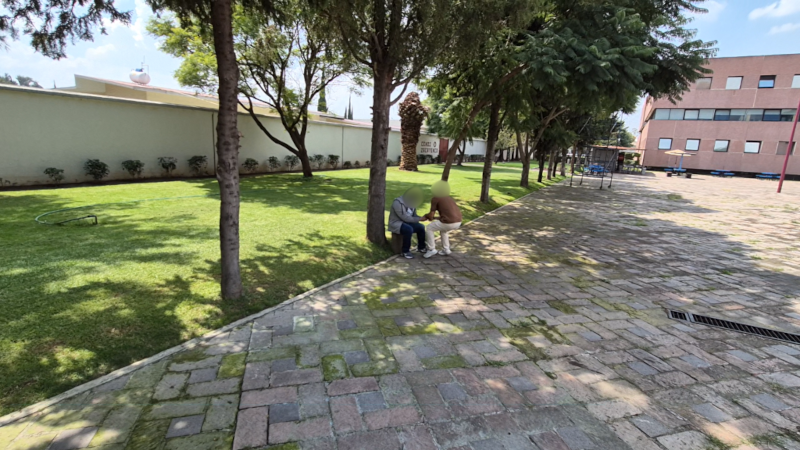}
        \includegraphics[width=0.24\linewidth]{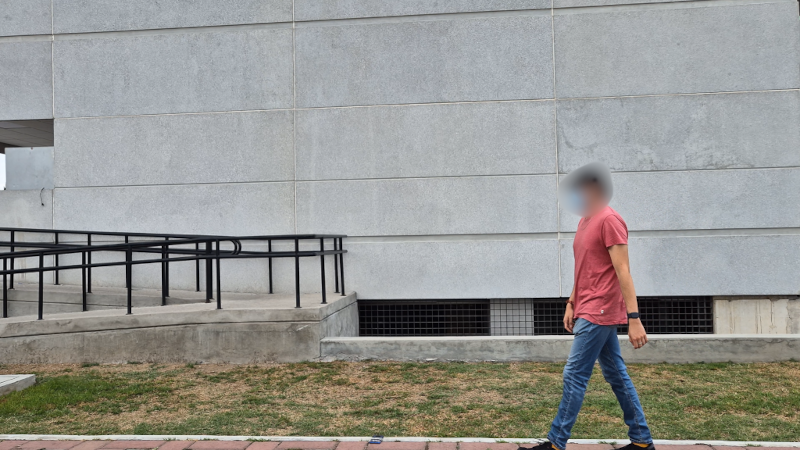}
        \includegraphics[width=0.24\linewidth]{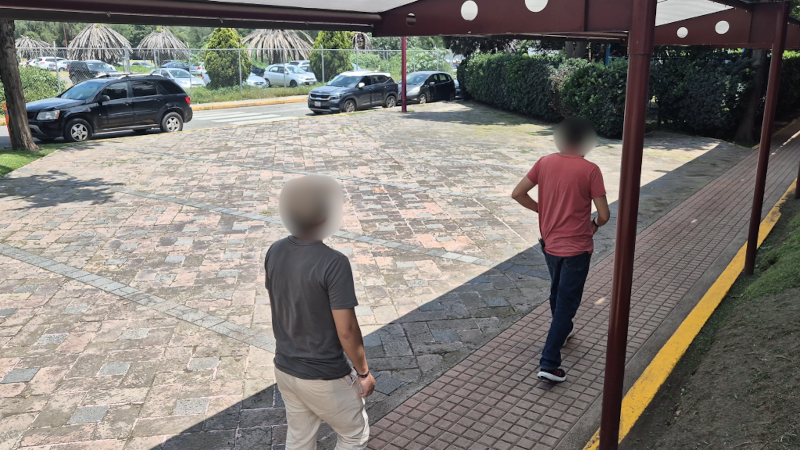}
        \includegraphics[width=0.24\linewidth]{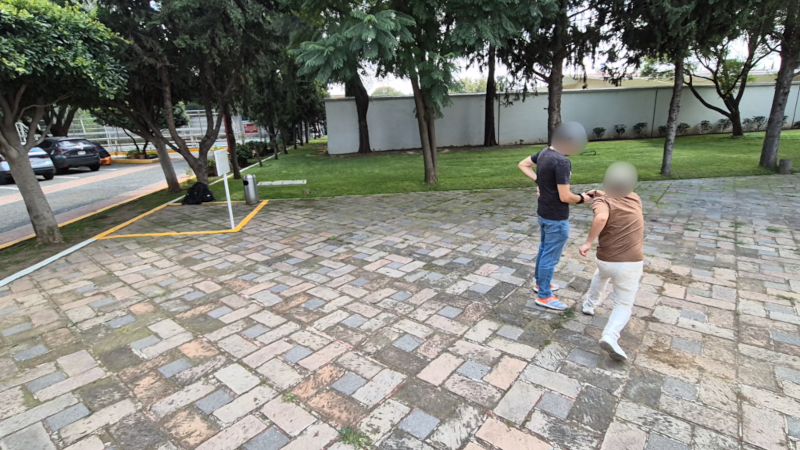}
    \caption{Representative frames from the training/validation video dataset recorded for this work. Faces were deliberately blurred for this paper.}
    \label{fig:train_dataset_frames}
\end{figure}

%\subsection{Dataset}
For train and validation, we created a dataset focused on non-violent robbery. We recorded staged snatching events across different days, times, viewpoints, and locations to increase contextual diversity while keeping the target action consistent.
The resulting dataset was labeled into two classes and then split into training and validation subsets. In total, it contains 90 examples: 29 robbery samples (positive class) and 61 non-robbery samples (negative class). Representative frames from the training/validation dataset are shown in Fig.~\ref{fig:train_dataset_frames}.

\begin{figure}[tb]
    \centering
    \includegraphics[width=0.24\textwidth]{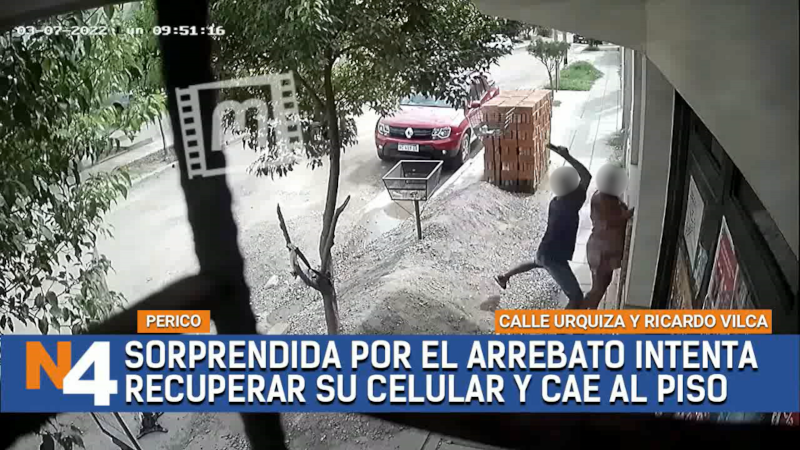}
    \includegraphics[width=0.24\textwidth]{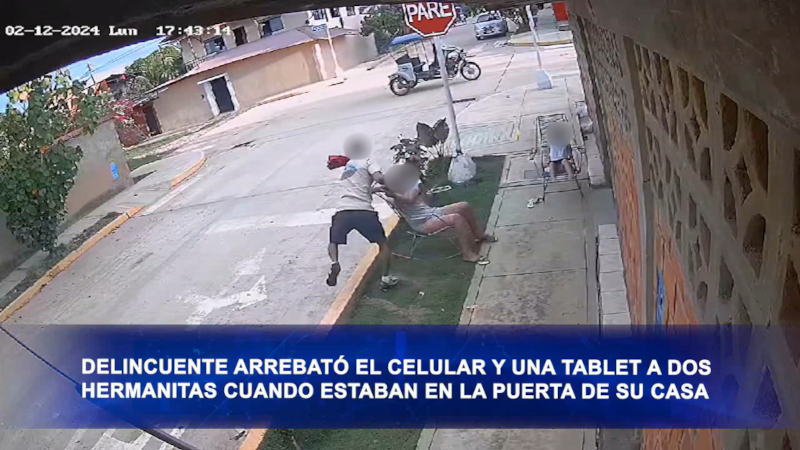}
    \includegraphics[width=0.24\textwidth]{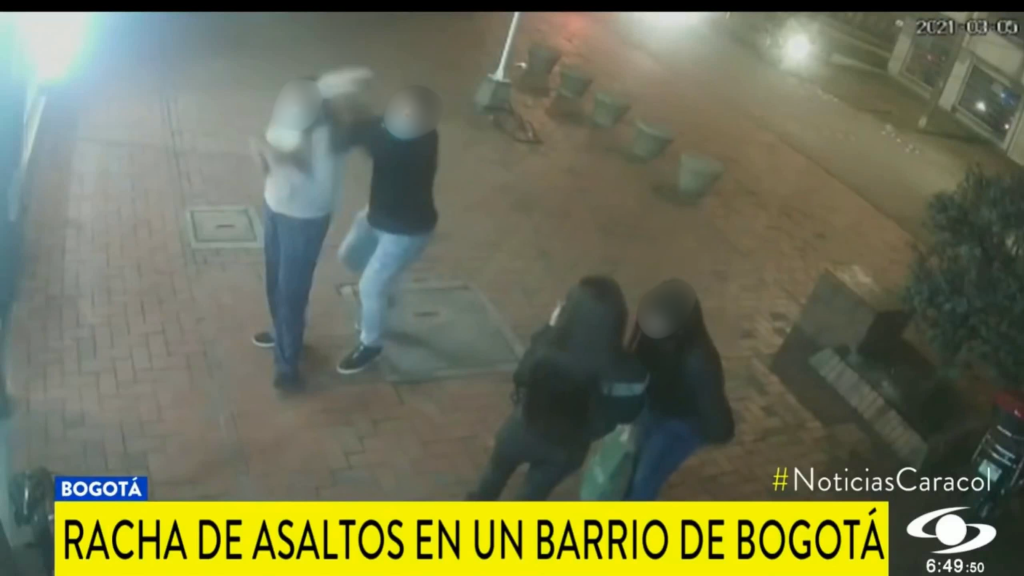}
    \includegraphics[width=0.24\textwidth]{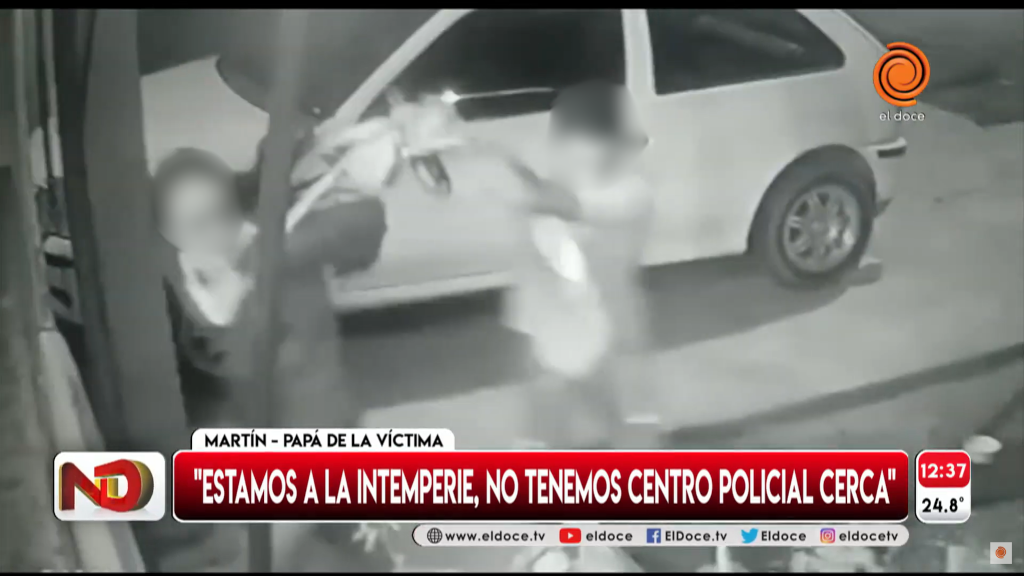}
    \label{fig:test_ex_4}
    \caption{Representative frames from the test video dataset. These videos are publicly available on YouTube. Faces were deliberately blurred for this paper.}
    \label{fig:test_dataset_frames}
\end{figure}

For testing, we collected a disjoint dataset from various sources on the internet. To better assess robustness, these clips include substantial contextual variability (e.g., different camera angles, different subjects, and variations in how the snatching event is executed). The labeled test set contains 47 examples: 17 robbery samples and 30 non-robbery samples. Representative frames from the test dataset are shown in Fig. \ref{fig:test_dataset_frames}.

\begin{figure*}
    \centering
    \includegraphics[width=0.19\linewidth]
    {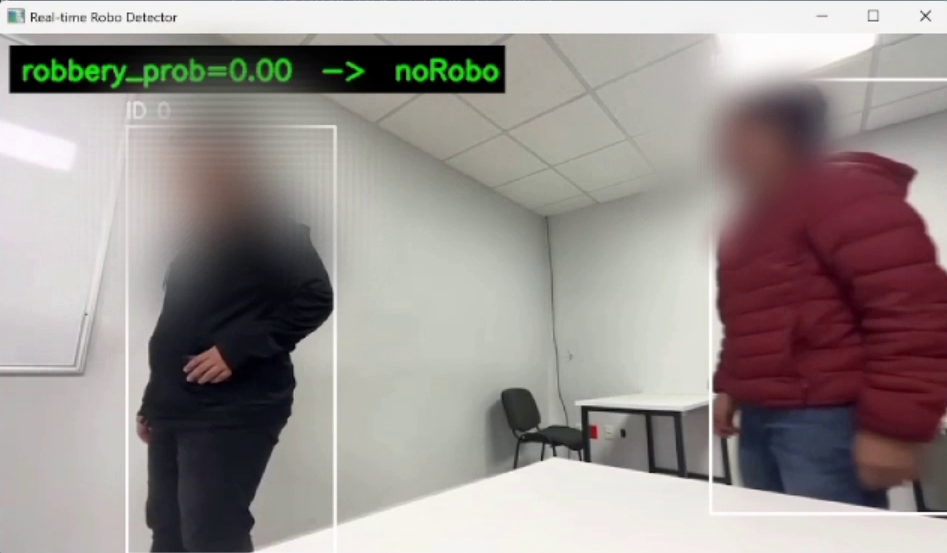}
    \includegraphics[width=0.19\linewidth]
    {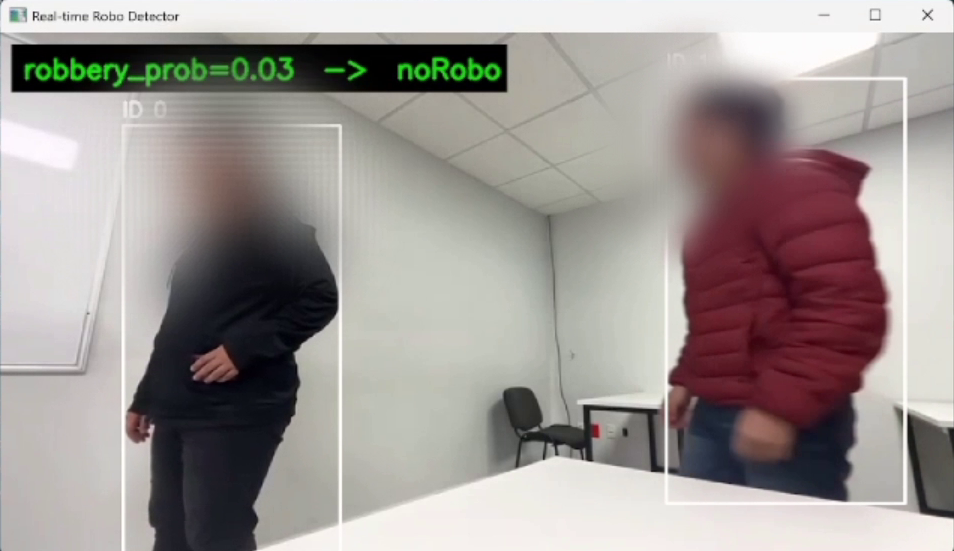}
    \includegraphics[width=0.19\linewidth]
    {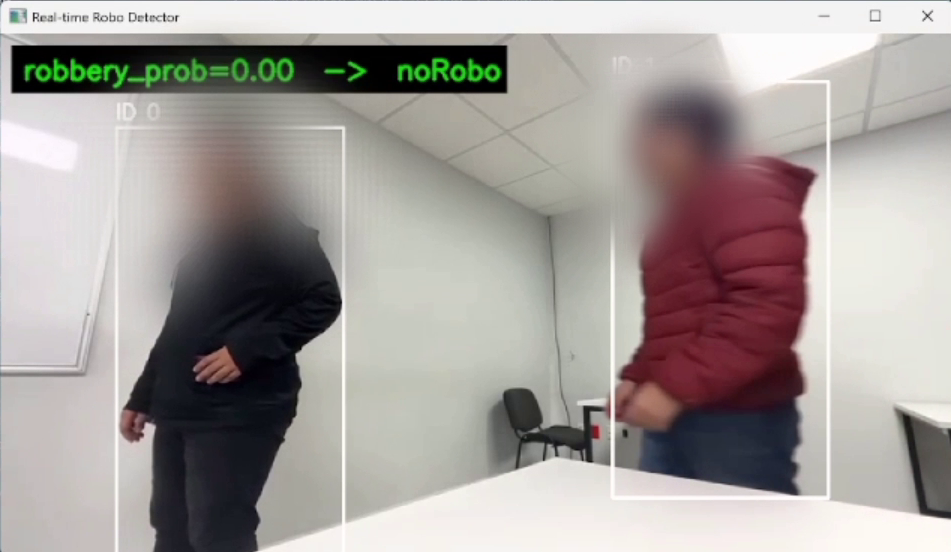}
    \includegraphics[width=0.19\linewidth]
    {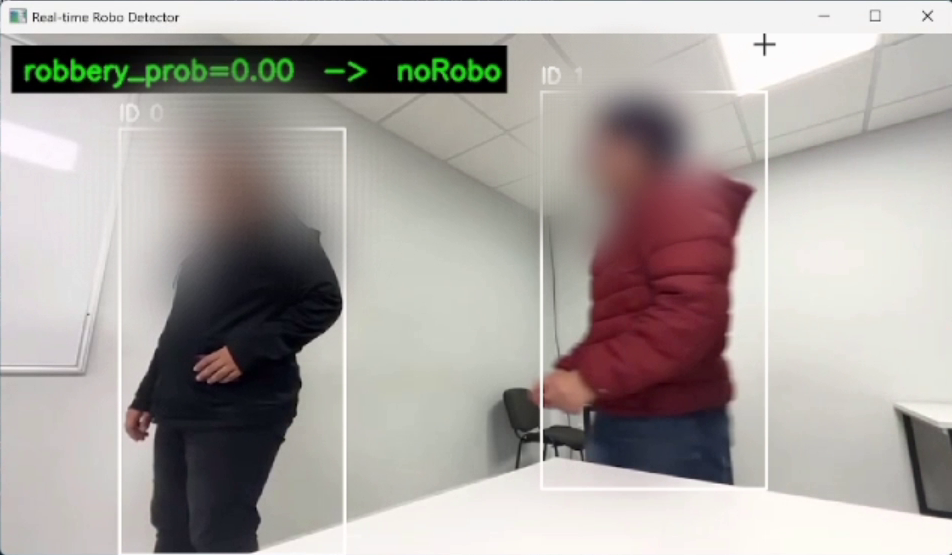}
    \includegraphics[width=0.19\linewidth]
    {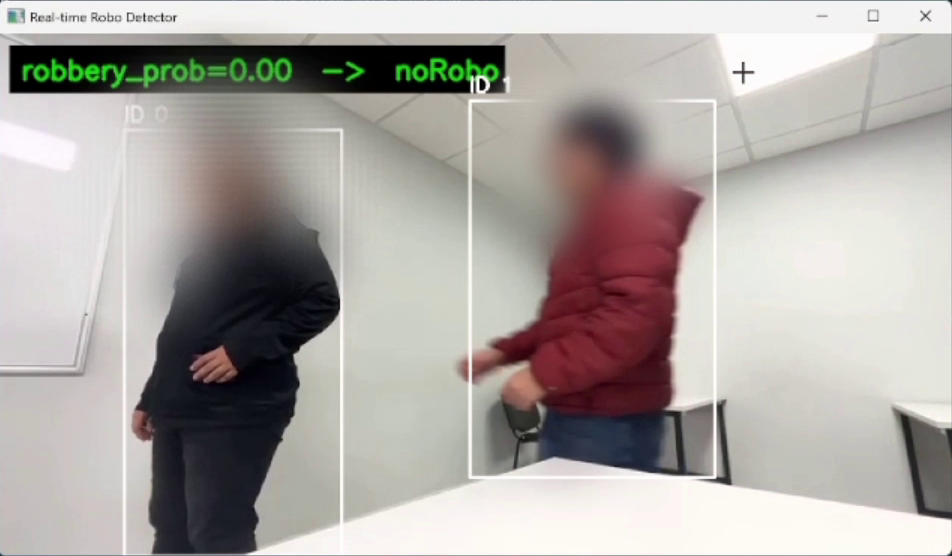}
    \includegraphics[width=0.19\linewidth]
    {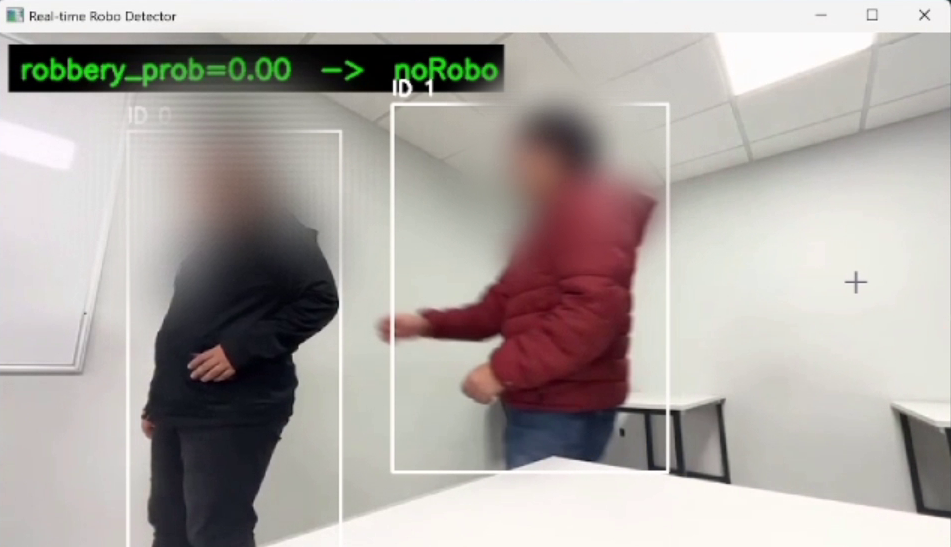}
    \includegraphics[width=0.19\linewidth]
    {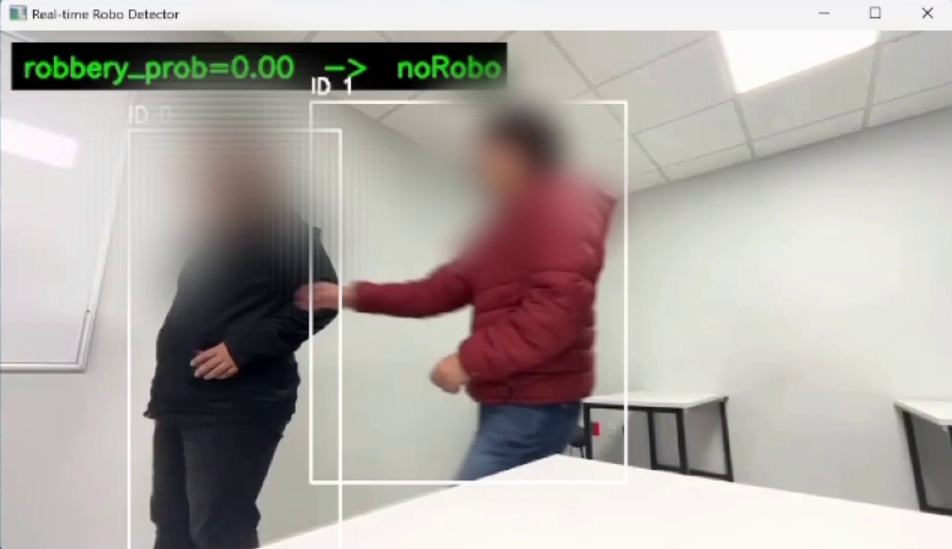}
    \includegraphics[width=0.19\linewidth]
    {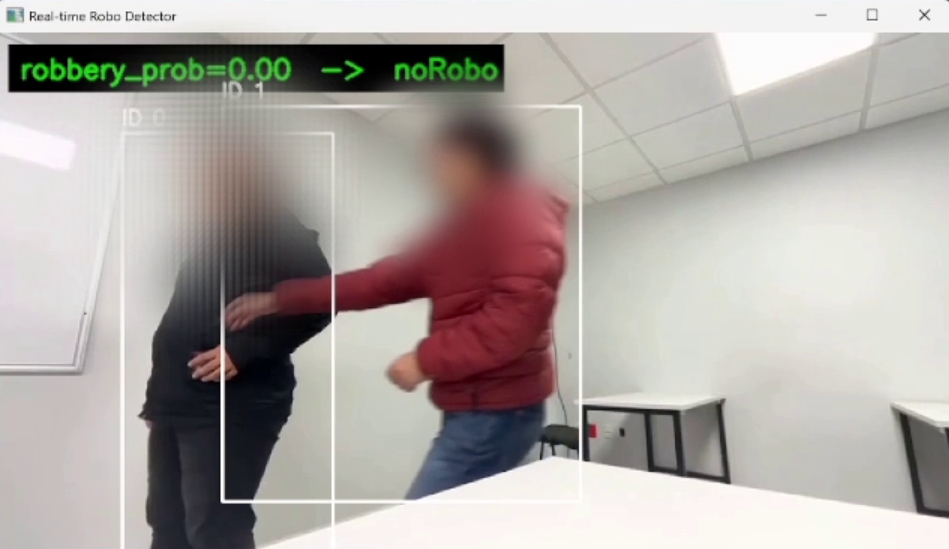}
    \includegraphics[width=0.19\linewidth]
    {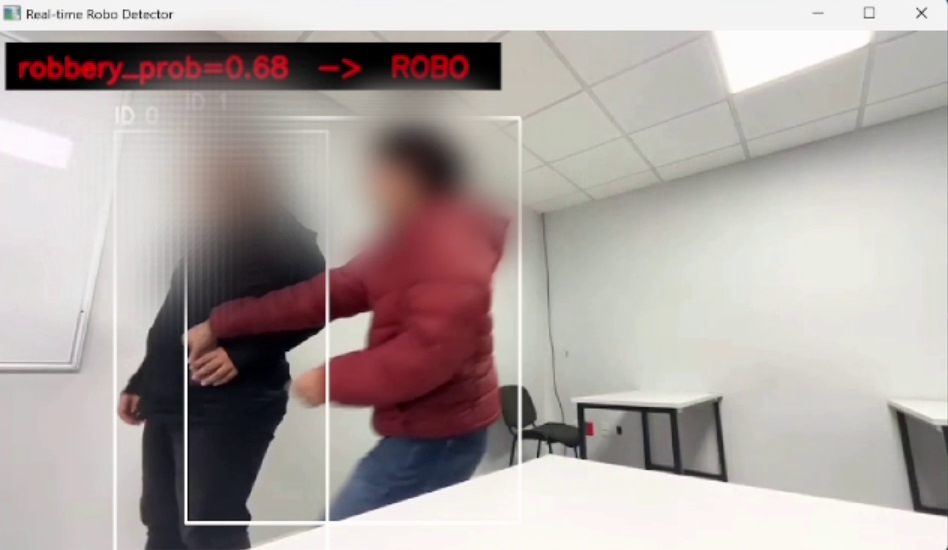}
    \includegraphics[width=0.19\linewidth]
    {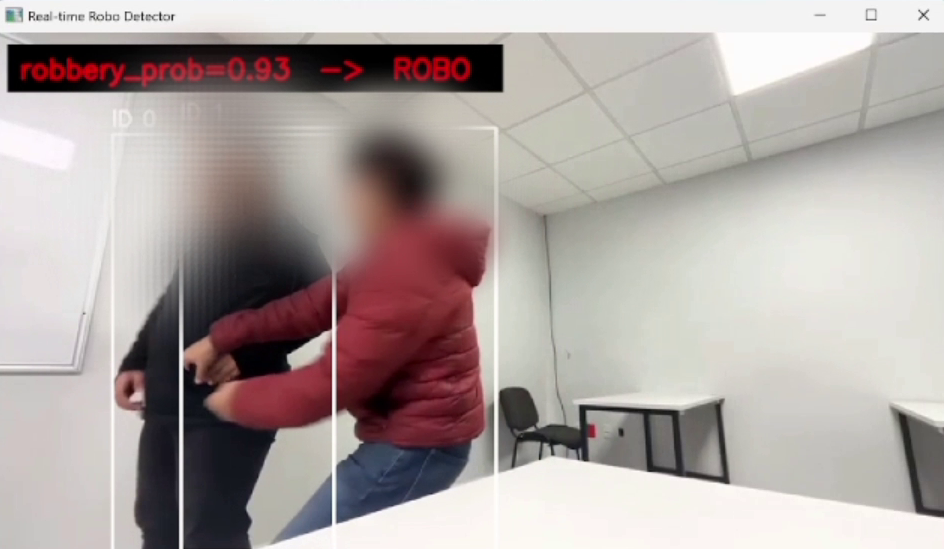}
    \caption{Example sequence of the functioning of the implemented method. The sequence shows the detection of a robbery.}
    \label{fig:extenden_example}
\end{figure*}

%\subsection{Model Training and Validation}

Classification was performed with a Random Forest model. To mitigate class imbalance, we used $n\_estimators = 500$, $random\_state = 42$, and $class\_weight = \text{'balanced'}$. The validation results are reported in Table~\ref{tab:val_report}.

The proposed method achieves an overall validation accuracy of $0.83$. For the non-robbery class, the model attains a precision of $0.91$ and an F1-score of $0.87$. Importantly, for the robbery class--the target event of interest--the model reaches a recall of $0.83$ and an F1-score of $0.77$, reflecting a favorable sensitivity to snatch-and-run behaviors while maintaining acceptable precision ($0.71$). These results support the capacity of the proposed pipeline to address the target task under realistic variability and limited data, providing a robust basis for proactive surveillance. Fig. \ref{fig:extenden_example} shows a sequence of frames of the implemented systems working on real time.

% --- Tabla 6.1 en LaTeX ---
\begin{table}[tb]
\centering
\caption{Classification results for validation stage.}
\label{tab:val_report}
\begin{tabular}{|l|c|c|c|c|c|}
\hline
\textbf{Class} & \textbf{Acc.} & \textbf{Prec.} & \textbf{Recall} & \textbf{F1-Score} & \textbf{Support} \\ \hline
Non-Robbery  & 0.83 & 0.91 & 0.83 & 0.87 & 12 \\
Robbery      & 0.83 & 0.71 & 0.83 & 0.77 & 6 \\ \hline
\end{tabular}
\end{table}

%During validation, the model achieved a precision of 0.91 for the "Non-Robbery" class, indicating a low rate of false negatives. For the "Robbery" class, the F1-Score of 0.77 demonstrates a stable balance between precision and sensitivity, which is vital for minimizing false alarms in a real-time security context.

%\subsection{Test Results}

The model was further evaluated on a held-out test set to assess generalization under higher contextual variability. As summarized in Table~\ref{tab:test_report}, the method achieves a class accuracy of $73.3\%$ for \emph{Non-Robbery} samples, with precision, recall and F1-score of $0.78$, $0.83$ and $0.81$.

For the target \emph{Robbery} class, the system obtains a precision of $0.67$, recall of $0.59$ and F1-score of $0.62$. While these values reflect the increased difficulty of recognizing subtle snatch-and-run events in unconstrained internet videos, the results still demonstrate that the pose-driven, feature-based methodology can detect a substantial portion of robberies while maintaining moderate precision. 

%This performance supports the feasibility of deploying the proposed approach for proactive surveillance, where the hysteresis filtering stage can further mitigate occasional frame-level fluctuations.

% --- Tablas 6.2 y 6.3 Combinadas ---
%\begin{table}[ht]
%\centering
%\caption{Confusion Matrix - Validation Phase (Raw and Normalized)}
%\label{tab:cm_val}
%\begin{tabular}{lcc|cc}
%\hline
% & \multicolumn{2}{c|}{\textbf{Raw Counts}} & \multicolumn{2}{c}{\textbf{Normalized}} \\ \hline
%\textbf{Real \textbackslash Pred} & \textbf{0} & \textbf{1} & \textbf{0 (\%)} & \textbf{1 (\%)} \\ \hline
%\textbf{Real 0} & 10 & 2 & 83.3\% & 16.7\% \\
%\textbf{Real 1} & 1 & 5 & 16.7\% & 83.3\% \\ \hline
%\end{tabular}
%\end{table}

% --- Tabla 6.4 en LaTeX ---
\begin{table}[tb]
\centering
\caption{Classification results for the test experiment.}
\label{tab:test_report}
\begin{tabular}{|l|c|c|c|c|c|}
\hline
\textbf{Class} & \textbf{Acc.} & \textbf{Precision} & \textbf{Recall} & \textbf{F1-Score} & \textbf{Support} \\ \hline
Non-Robbery & 73.3 & 0.78 & 0.83 & 0.81 & 30 \\
Robbery     & 58.8 & 0.67 & 0.59 & 0.62 & 17 \\ \hline
\end{tabular}
\end{table}

%% --- Tablas 6.5 y 6.6 Combinadas ---
%\begin{table}[ht]
%\centering
%\caption{Confusion Matrix - Test Phase (Raw and Normalized)}
%\label{tab:cm_test}
%\begin{tabular}{lcc|cc}
%\hline
% & \multicolumn{2}{c|}{\textbf{Raw Counts}} & \multicolumn{2}{c}{\textbf{Normalized}} \\ \hline
%\textbf{Real \textbackslash Pred} & \textbf{0} & \textbf{1} & \textbf{0 (\%)} & \textbf{1 (\%)} \\ \hline
%\textbf{Real 0} & 22 & 8 & 73.3\% & 26.6\% \\
%\textbf{Real 1} & 7 & 10 & 41.1\% & 58.8\% \\ \hline
%\end{tabular}
%\end{table}

%\subsection{Discussion}

The experimental data reveals that the system performs adequately even for the test set that uses quite different scenes. The occasional misclassifications observed in the test phase can be attributed to camera angles, lighting variations, and the inherent diversity of "snatch-and-run" movements in low-resolution internet footage. Nevertheless, the results on the internet-based test set provide a benchmark for this variability. This performance reflects a deliberate design trade-off that prioritizes interpretability and real-time execution. The current effectiveness of the method is centered on rapid interactions between two individuals in scenarios with adequate visibility and lighting. Additionally, the focus on binary interactions is consistent with the nature of snatching events, where the interaction typically involves two individuals in close proximity. Despite these challenges, the use of a lightweight YOLO model combined with a Random Forest classifier allowed for an adequate response time on the NVIDIA Jetson Nano, validating the feasibility of the proposed architecture for real-time proactive surveillance.

\section{Conclusions}
\label{sec:conclu}

This study presented a method for detecting non-violent robbery (snatch-and-run) in surveillance videos. Starting from raw video, the pipeline combines pose keypoints and an interpretable, feature-based classifier to recognize suspicious interactions. To improve temporal robustness, we incorporated a hysteresis filtering stage that reduces sporadic false positives in frame-level predictions. While the current findings are based on a limited dataset, the experimental results show promising performance across videos captured in different contexts. These results suggest the potential of the proposed approach for real-world surveillance. Finally, we validated the implementation on an NVIDIA Jetson device, demonstrating that the method can run on edge hardware. Future work will focus on addressing class imbalance and extending the model to additional behaviors.

\begin{credits}
%\subsubsection{\ackname} This study was funded by Secretaría de Investigación y Posgrado (SIP), Instituto Politécnico Nacional, Research Projects 20250144 and 20250342.

%\subsubsection{\discintname}
%The authors have no competing interests to declare that arerelevant to the content of this article. 
\end{credits}
\bibliographystyle{splncs04}
\bibliography{referencias}
\end{document}